\documentclass{article}

\usepackage{PRIMEarxiv}

\usepackage[utf8]{inputenc} 
\usepackage[T1]{fontenc}    
\usepackage{hyperref}       
\usepackage{url}            
\usepackage{booktabs}       
\usepackage{amsfonts}       
\usepackage{nicefrac}       
\usepackage{microtype}      
\usepackage{lipsum}
\usepackage{fancyhdr}       
\usepackage{graphicx}       
\graphicspath{{media/}}     
\usepackage{multirow}
\title{WaveMixSR: A Resource-efficient Neural Network for Image Super-resolution
}

\author{
  Pranav Jeevan, Akella Srinidhi, Pasunuri Prathiba, Amit Sethi \\
  Department of Electrical Engineering \\
  Indian Institute of Technology Bombay \\
  Mumbai, India\\
  \texttt{\{194070025, 213079003, asethi \}@iitb.ac.in} \\
   \And
}

\begin{document}
\maketitle

\begin{abstract}
Image super-resolution research recently been dominated by transformer models which need higher computational resources than CNNs due to the quadratic complexity of self-attention. We propose a new neural network -- WaveMixSR -- for image super-resolution based on WaveMix architecture which uses a 2D-discrete wavelet transform for spatial token-mixing. Unlike transformer-based models, WaveMixSR does not unroll the image as a sequence of pixels/patches. It uses the inductive bias of convolutions along with the lossless token-mixing property of wavelet transform to achieve higher performance while requiring fewer resources and training data. We compare the performance of our network with other state-of-the-art methods for image super-resolution. Our experiments show that WaveMixSR achieves competitive performance in all datasets and reaches state-of-the-art performance in the BSD100 dataset on multiple super-resolution tasks. Our model is able to achieve this performance using less training data and computational resources while maintaining high parameter efficiency compared to current state-of-the-art models.
\end{abstract}


\section{Introduction}
\label{sec:intro}

\begin{figure}[b]

\centering
\includegraphics[scale=0.66]{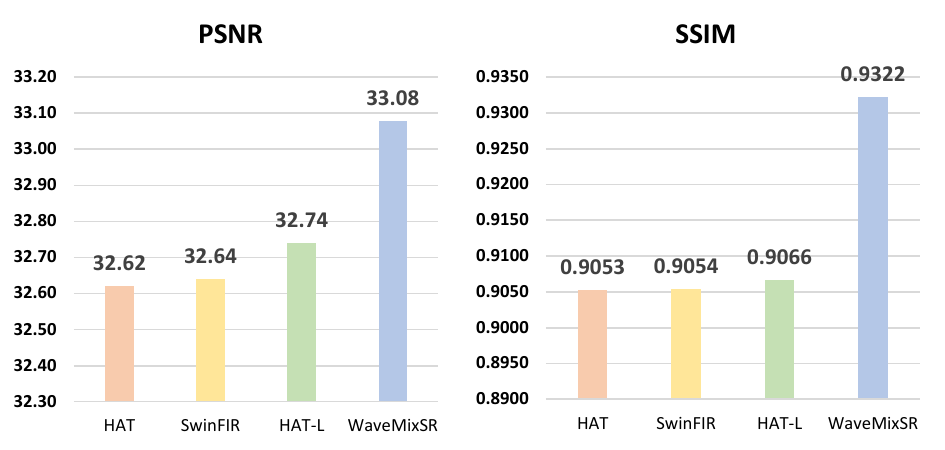}
\caption{Comparison of PSNR and SSIM for $2\times$ SR on BSD100 dataset shows WaveMixSR surpasses the previous state-of-the-art methods such as HAT~\cite{Chen_2023_CVPR} and SwinFIR~\cite{zhang2023swinfir}.}
\label{fig:graph}

\end{figure}

Single image super-resolution (SR) aims to enhance the resolution of single images, which makes it an ill-posed problem. This task has various applications in fields such as medical imaging, remote sensing, digital photography, and surveillance. Image SR is a challenging task due to the lack of information to completely reconstruct the high-resolution output. It is a low-level vision task along with other single image reconstruction tasks, such as image denoising and JPEG image artifact removal. 

In recent years, various deep learning techniques have shown remarkable progress in image super-resolution and have achieved state-of-the-art performance in terms of both quantitative metrics and visual quality. During the initial application of deep learning to SR, convolutional neural network (CNN) based methods were used widely. Recently, after the widespread success of attention-based transformer~\cite{vaswani2017attention} models in natural language processing tasks, they have also been used for computer vision tasks. Vision transformer and its variants have been outperforming CNNs in most of the vision tasks, including SR. Transformer-based models such as Swin transformers have been used in both high-level and low-level vision tasks. The recently developed SwinIR~\cite{liang2021swinir} and Hybrid Attention Transformer (HAT)~\cite{Chen_2023_CVPR} have outperformed the previous convolutional models in SR. The success of transformer-based models showed that long-range spatial token-mixing by self-attention can outperform convolutional networks in super-resolution tasks.

The application of transformers in computer vision has been limited by the quadratic complexity of self-attention with respect to the input sequence length. Images, when unrolled into long sequences impose a large computational and memory burden on GPUs. Moreover, transformers do not have an inductive bias unlike CNNs. Therefore they require a large number of training images to give good generalization compared to the convolutional models. Vision transformers also needs extra architectural changes, such as the inclusion of positional information through learned or fixed positional embeddings. This demand for a large amount of data and GPU resources is not suitable for resource-constrained scenarios where data and GPU are limited, such as in green or edge computing.

On the other hand, CNNs have inductive priors, such as translational equivariance due to convolutional weight sharing and partial scale invariance due to pooling, to handle 2D images which enables them to learn from smaller datasets with less computational expenditure. But, they fail to capture long-range dependencies compared to transformers and several layers to increase their receptive fields. 

Intuitively, SR is a low-level vision task that needs more local information than global context for generating HR images from LR images. The better performance of transformer models compared to CNN can perhaps be attributed to the token-mixing ability of self-attention rather than modeling long-range interactions.


Recently, new token-mixing models have emerged, replacing the self-attention in transformers with other mechanisms such as Fourier transforms, depth-wise convolutions, wavelet transforms, and spatial multi-layer perceptrons. Their models have been shown to perform on par with self-attention-based transformer models without suffering the large data and GPU requirements. For instance, the WaveMix~\cite{jeevan2023wavemix} architecture which uses 2-Dimensional discrete wavelet transform (2D DWT) for token-mixing has been shown to perform well in high-level vision tasks such as image classification and semantic segmentation while consuming low computational resources. Wavemix uses self-similar WaveMix blocks to process images and is found to be versatile as the same model can be used for different tasks by just changing the final output layer. The performance of 2D DWT token-mixing for image SR has not been tested. 

We use the WaveMix blocks introduced in ~\cite{jeevan2023wavemix} to create a new token-mixing architecture for image SR. Equipped with spatial token-mixing using discrete wavelet transform and the inductive bias of CNNs, WaveMixSR can surpass the state-of-the-art methods in SR of the BSD100 dataset as shown in Table~\ref{tab:bsd100}. The contributions in this paper are: 
\begin{itemize}
    \item We propose a new network that employs spatial token-mixing using 2D discrete wavelet transform for single image super-resolution.
    \item Our model is efficient in terms of computation resource utilization and parameter count. WaveMixSR consumes less than half the parameters and resources compared to other state-of-the-art models.
    \item WaveMixSR achieves state-of-the-art performance in multiple SR tasks on the BSD100 dataset. It achieves this performance without the need for large training data and pre-training procedures employed in other models.
\end{itemize}

\section{Related Works}

\subsection{CNN-based methods}
SRCNN~\cite{dong2015image} was one of the first deep learning models developed for single image SR to directly learn an end-to-end mapping between the low and high-resolution images. Enhanced Deep Residual Networks~\cite{lim2017enhanced} (EDSR) was an improvement over SRCNN which produced refined results by using residual blocks and removed unnecessary modules to stabilize training. Residual Channel Attention Networks~\cite{zhang2018image} (RCAN) uses a channel attention mechanism that adaptively recalibrates channel-wise features by explicitly modeling inter-dependencies between channels. It also employs a residual in residual (RIR) structure to form deep networks that have long and short skip connections for better gradient flow. Holistic Attention Network~\cite{niu2020single} (HAN) enhances the representation of the image features using a layer attention module (LAM) and a channel-spatial attention module (CSAM). LAM adaptively emphasizes hierarchical features by considering correlations among layers while CSAM learns the confidence at all the positions of each channel to selectively capture more informative features. The CNN-based models were state-of-the-art till transformer-based models started being used for computer vision tasks after the coming of vision transformers~\cite{dosovitskiy2021image}.

\subsection{Transformer-based methods}

After the success of transformers in Natural language processing, they have been used for computer vision tasks with the advent of vision transformers~\cite{dosovitskiy2021image}. They have been outperforming CNNs in classification, segmentation, and detection tasks, especially when large datasets and compute are available. This has led to the application of transformer-based models in low-level vision tasks such as SR. Image processing transformer~\cite{chen2021pretrained} (IPT) introduced multi-task pre-training on a vision transformer model for low-level vision tasks but suffered from a large number of parameters (116 M) and computational costs.  Encoder-decoder-based transformer~\cite{li2022efficient} (EDT) tried to overcome the large computational requirements of IPT by employing multi-related-task pre-training.

SwinIR~\cite{liang2021swinir} used the Swin transformer for deep feature extraction and outperformed CNN-based models in multiple image restoration tasks such as SR, denoising, and compression artifact reduction.  SwinIR consists of shallow and deep feature extraction modules composed of Swin transformer layers and a high-quality image reconstruction module. SwinFIR~\cite{zhang2023swinfir} improves upon SwinIR by using Fast Fourier Convolution (FFC) components which improves the efficiency in capturing global information due to its image-wide receptive field. It also employed advanced data augmentations, pre-training, and feature ensemble to improve image reconstruction.

HAT~\cite{Chen_2023_CVPR} combined self-attention, channel attention, and overlapping cross-attention for giving the state-of-the-art performance in SR across various benchmark datasets.

\subsection{Wavelet-based methods}

Recognizing that wavelet transform provides an \emph{approximation} as well as \emph{detail} information of an image, deep wavelet super resolution~\cite{8014882} (DWSR) used a deep CNN to reconstruct the wavelet coefficients of LR images to obtain the corresponding SR outputs. They showed that wavelets provided an image representation that simplifies the mapping between LR and HR images, aiding in faster image reconstruction. ~\cite{kumar2017convolutional} uses a CNN to estimate wavelet detail coefficients of a desired high resolution (HR) image  on the given low resolution (LR) image.  

Wavelet-based super-resolution net~\cite{8237449} (Wavelet-SRNet) was developed for human face SR. It predicted the wavelet coefficients of HR images before reconstructing HR images from them. Different loss functions were also employed to capture both global topology information and local texture details of human faces.

Multi-level wavelet CNN~\cite{liu2019multilevel} (MWCNN) was proposed to increase receptive field size and computational efficiency by replacing pooling with wavelet transform to reduce the resolution of feature maps. MWCNN for image restoration tasks was based on the U-Net architecture and used inverse wavelet transform (IWT) to reconstruct the high-resolution (HR) feature maps. 

\begin{figure*}[ht]

\centering
\includegraphics[scale=0.85]{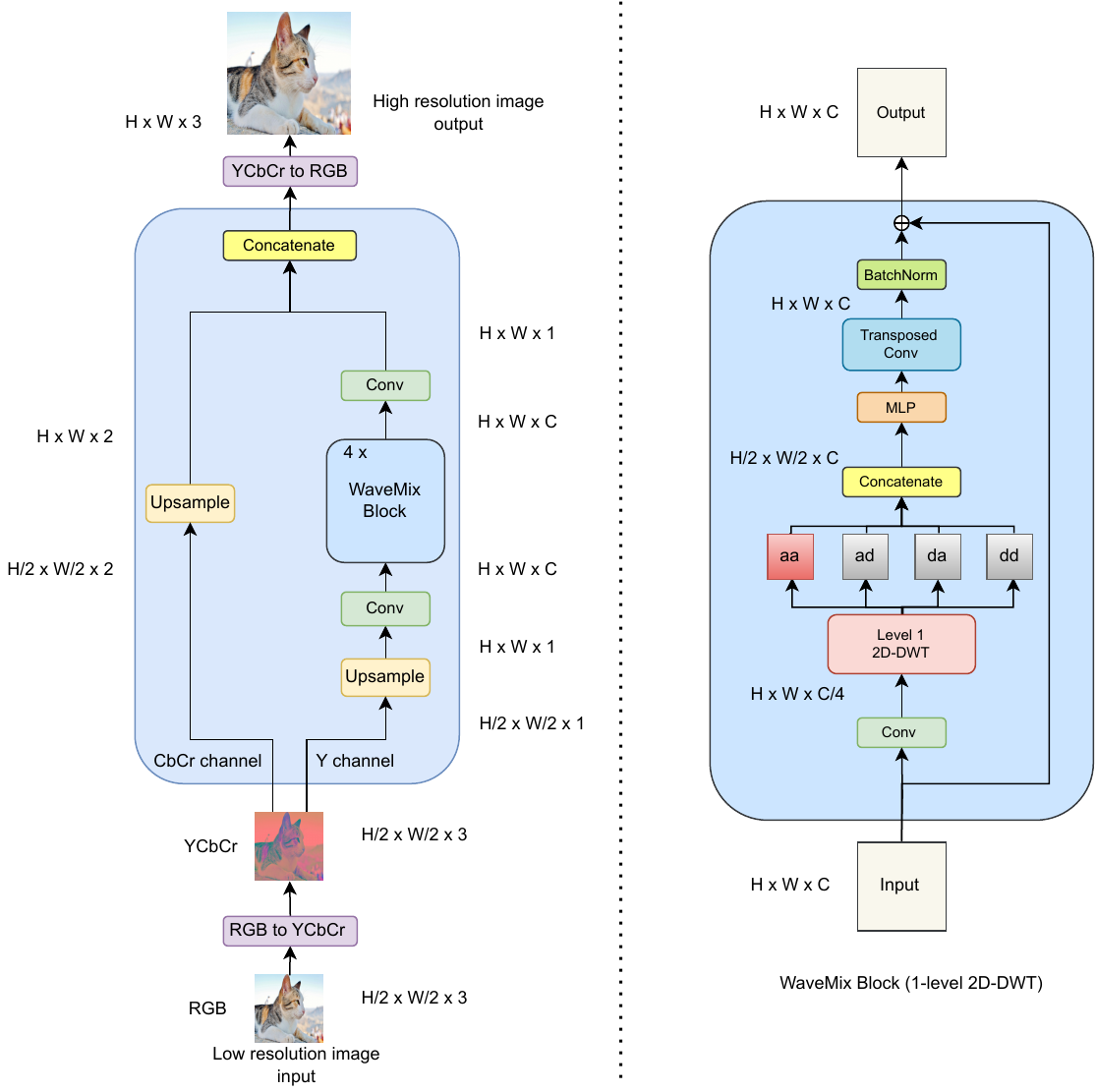}
\caption{Architecture of WaveMixSR along with the details of the WaveMix block. Architecture is shown for $2\times$ super-resolution. For $3\times$  and $4\times$ super-resolution, we only modify the upsample blocks in the architecture keeping the other blocks intact.}
\label{fig:WaveMixSR}

\end{figure*}
The invertibility of wavelet transforms ensures that none of the image information or features are lost during the down-sampling operation.

\section{Architecture}

As shown in Figure~\ref{fig:WaveMixSR}, the network consists of two paths for generating the HR image from the LR image. Feature map expansion of the input is done primarily by the non-parametric upsampling operations. Model focuses on sharpening the up-sampled images by reconstructing details which are missing. 

\subsection{WaveMixSR}

The LR input image in RGB space is converted to YCbCr space before sending to the model. The YCbCr color space separates the image into three channels: Y (luma), Cb (blue-difference chroma), and Cr (red-difference chroma). The Y channel represents the luminance information of the image, while Cb and Cr channels represent the chrominance information. WaveMixSR model has two paths -- one for handling the Y channel and another for the CbCr channels of the input image. Typically, in SR techniques only the Y channel is used for the path with parametric learning because the Y channel contains most of the image details and is less affected by color changes. This approach has been shown to improve the performance of deep learning-based SR methods~\cite{dong2015image}. We follow the same idea in WaveMixSR, where only the Y channel is sent to the path with WaveMix blocks to focus on improving the resolution of the LR image while ignoring the color changes.

For $2\times$ SR, the first path takes the Y-channel feature map, $\textbf{y}\in\mathbb{R}^{H/2\times W/2 \times 1}$ of the input image and sends it through a parameter-free upsampling layer which resizes the low-resolution Y-channel image to high resolution using bilinear or bicubic interpolation. For $3\times $ and $ 4\times$ SR, we use corresponding upsampling blocks that upsample the image to HR. The output of upsampling block, $\textbf{y}\in\mathbb{R}^{H\times W \times 1}$ is sent to a convolutional layer to increase the number of feature maps from $1$ to $C$. The output feature maps from the convolutional layer, $\textbf{x}_{in}\in\mathbb{R}^{H\times W \times C}$ is sent to the WaveMix blocks. Four WaveMix blocks are connected in series in our model to create high-resolution feature maps. The output $\textbf{x}_{out}\in\mathbb{R}^{H\times W \times C}$ from the final WaveMix block is then passed through a convolutional layer which reduces the channel dimension $C$ and returns a single channel output $\textbf{y}_{out}\in\mathbb{R}^{H\times W \times 1}$. 

The second parallel path in WaveMixSR takes the two channels of CbCr and simply passes it through an upsampling layer where the channel resolution is increased to the final resolution required. This HR CbCr channel is finally concatenated with the Y-channel output $\textbf{y}_{out}$ from the first path, thereby creating the 3-channel YCbCr output. This output is then converted to an RGB color space to obtain the final high-resolution output image.

\subsection{WaveMix block}

We use the WaveMix block with a single level of 2D-discrete wavelet transform~\cite{jeevan2023wavemix}. As shown in Figure~\ref{fig:WaveMixSR}, the design of the WaveMix block is such that it does not collapse the spatial resolution of the feature maps, unlike CNN blocks that use pooling operations~\cite{https://doi.org/10.48550/arxiv.1512.03385}. And yet, it reduces the number of computations required by reducing the spatial dimensions of the feature maps using 2D-DWT, which translates to a reduction in GPU RAM, training time, and inference time. However, unlike pooling or strided convolutions, a 2D-DWT is lossless as it expands the number of channels by the same factor by which it reduces spatial resolution. Furthermore, it has additional energy compaction (sparsification) properties that are not offered by random filters or Fourier basis.

Denoting input and output tensors of the WaveMix block by $\textbf{x}_{in}$ and $\textbf{x}_{out}$, respectively; the four wavelet filters along with their downsampling operations at each level by $w_{aa},w_{ad},w_{da},w_{dd}$ ($a$ for approximation, $d$ for detail); convolution, multi-layer perceptron (MLP), transposed convolution (upconvolution), and batch normalization operations by $c$, $m$, $t$, and $b$, respectively; and their respective trainable parameter sets by $\xi$, $\theta$, $\phi$, and $\gamma$, respectively; concatenation along the channel dimension by $\oplus$, and point-wise addition by $+$, the operations inside a WaveMix block can be expressed using the following equations:

\begin{equation} \label{eq:1}
    \textbf{x}_0 = c(\textbf{x}_{in},\xi);  \hspace{2cm}    
    \textbf{x}_{in}\in\mathbb{R}^{H\times W \times C}, \textbf{x}_0\in\mathbb{R}^{H\times W \times C/4}    
\end{equation}

\begin{equation}\label{eq:2}
    \textbf{x} = [w_{aa}(\textbf{x}_0) \oplus w_{ad}(\textbf{x}_0) \oplus w_{da}(\textbf{x}_0) \oplus w_{dd}(\textbf{x}_0)]; \hspace{1cm}
    \textbf{x}\in\mathbb{R}^{H/2\times W/2 \times 4C/4}
\end{equation}

\begin{equation} \label{eq:3}
    \tilde{\textbf{x}} = b(t(m(\textbf{x},\theta),\phi),\gamma); \hspace{3cm}\tilde{\textbf{x}}\in\mathbb{R}^{H\times W \times C}      
\end{equation}

\begin{equation}\label{eq:4}
    \textbf{x}_{out} = \tilde{\textbf{x}}_1 + \textbf{x}_{in}; \hspace{3cm}\textbf{x}_{out} \in\mathbb{R}^{H\times W \times C} 
\end{equation}

\begin{table*}[]
\centering
\begin{tabular}{@{}llrrllrr@{}}
\toprule
\multirow{2}{*}{Model} & \multirow{2}{*}{Training dataset} & \multicolumn{6}{c}{Testing metrics on BSD100} \\ \cmidrule(l){3-8} 
 &  & \multicolumn{2}{c}{$2\times$} & \multicolumn{2}{c}{$3\times$} & \multicolumn{2}{c}{$4\times$} \\ \cmidrule(l){3-8} 
 &  & PSNR & SSIM & PSNR & SSIM & PSNR & SSIM \\ \midrule
EDSR~\cite{lim2017enhanced} & DIV2K & 32.32 & 0.9013 & 29.25 & 0.8093 & 27.71 & 0.7420 \\
RCAN~\cite{zhang2018image} & DIV2K & 32.41 & 0.9027 & 29.32 & 0.8111 & 27.77 & 0.7436 \\
SAN~\cite{Chen_2023_CVPR} & DIV2K & 32.42 & 0.9028 & 29.33 & 0.8112 & 27.78 & 0.7436 \\
IGNN~\cite{Chen_2023_CVPR} & DIV2K & 32.41 & 0.9025 & 29.31 & 0.8105 & 27.77 & 0.7434 \\
HAN~\cite{niu2020single} & DIV2K & 32.41 & 0.9027 & 29.32 & 0.8110 & 27.80 & 0.7442 \\
NLSN~\cite{Chen_2023_CVPR} & DIV2K & 32.43 & 0.9027 & 29.34 & 0.8117 & 27.78 & 0.7444 \\
RCN-it~\cite{Chen_2023_CVPR} & DF2K & 32.48 & 0.9034 & 29.39 & 0.8125 & 27.87 & 0.7459 \\
SwinIR~\cite{liang2021swinir} & DF2K & 32.53 & 0.9041 & 29.46 & 0.8145 & 27.92 & 0.7489 \\
EDT~\cite{li2022efficient} & DF2K & 32.52 & 0.9041 & 29.44 & 0.8142 & 27.91 & 0.7483 \\
HAT~\cite{Chen_2023_CVPR} & DF2K & 32.62 & 0.9053 & 29.54 & 0.8167 & 28.00 & 0.7517 \\
SwinFIR*~\cite{zhang2023swinfir} & DF2K & 32.64 & 0.9054 & 29.55 & 0.8169 & 28.03& 0.7520 \\
HAT-L*~\cite{Chen_2023_CVPR} & DF2K & 32.74 & 0.9066 & \textbf{29.63} & \textbf{0.8191} & \textbf{28.09} & 0.7551 \\
WaveMixSR & DIV2K & \textbf{33.08} & \textbf{0.9322} & 28.38 & 0.8043 & 27.65 & \textbf{0.7605} \\ \bottomrule
\end{tabular}
\caption{Quantitative comparison with state-of-the-art methods on the BSD100 dataset shows that WaveMixSR performs better using less training data (* indicates models that were pre-trained on ImageNet).}
\label{tab:bsd100}
\end{table*}

The WaveMix block extracts learnable and space-invariant features using a convolutional layer, followed by spatial token-mixing and downsampling for scale-invariant feature extraction using 2D-DWT~\cite{pyTorchWavelets}, followed by channel-mixing using a learnable MLP (1$\times$1 conv) layer, followed by restoring spatial resolution of the feature map using a transposed-convolutional layer. The use of trainable convolutions \emph{before} the wavelet transform allows the extraction of only those feature maps that are suitable for the chosen wavelet basis functions. The convolutional layer $c$ decreases the embedding dimension  $C$ by a factor of four so that the concatenated output $\textbf{x}$ after 2D-DWT has the same number of channels as the input $\textbf{x}_{in}$ (Eq.~\ref{eq:1} and Eq.2\ref{eq:2}). That is since 2D-DWT is a lossless transform, it expands the number of channels by the same factor (using concatenation) by which it reduces the spatial resolution by computing an approximation sub-band (low-resolution approximation) and three detail sub-bands (spatial derivatives)~\cite{57199} for each input channel (Eq.2\ref{eq:2}). The use of this image-appropriate and lossless downsampling using 2D-DWT allows WaveMix to use fewer layers and parameters.

The output $\hat{\textbf{x}}$ is then passed to an MLP layer $m$, which has two $1\times1$ convolutional layers with an inverse bottleneck design (multiplication factor $> 1$) separated by a GELU non-linearity. After this, the feature map resolution is doubled using transposed-convolutions $t$ followed by batch normalization $b$ (Eq.~\ref{eq:3}). A residual connection is used to ease the flow of the gradient~\cite{he2015deep} (Eq.4~\ref{eq:4}). 

Among the different types of mother wavelets available, we used the Haar wavelet (a special case of the Daubechies wavelet~\cite{57199}, also known as Db1), which is frequently used due to its simplicity and faster computation. Haar wavelet is both orthogonal and symmetric in nature and has been extensively used to extract basic structural information from images~\cite{Porwik2004TheHT}. For even-sized images, it reduces the dimensions exactly by a factor of $2$, which simplifies the designing of the subsequent layers.

Since the upsampling is done using parameter-free interpolation techniques like bilinear and bicubic interpolation in Pytorch, WaveMix blocks can focus on reconstructing the HR details and sharpening of the images. The same network can be used for performing any SR task ($2\times, 3\times $ or $ 4\times)$ by only modifying the upsampling blocks without any architectural changes or increasing the number of parameters.

\begin{figure}[!b]
\centering
\includegraphics[scale=0.5]{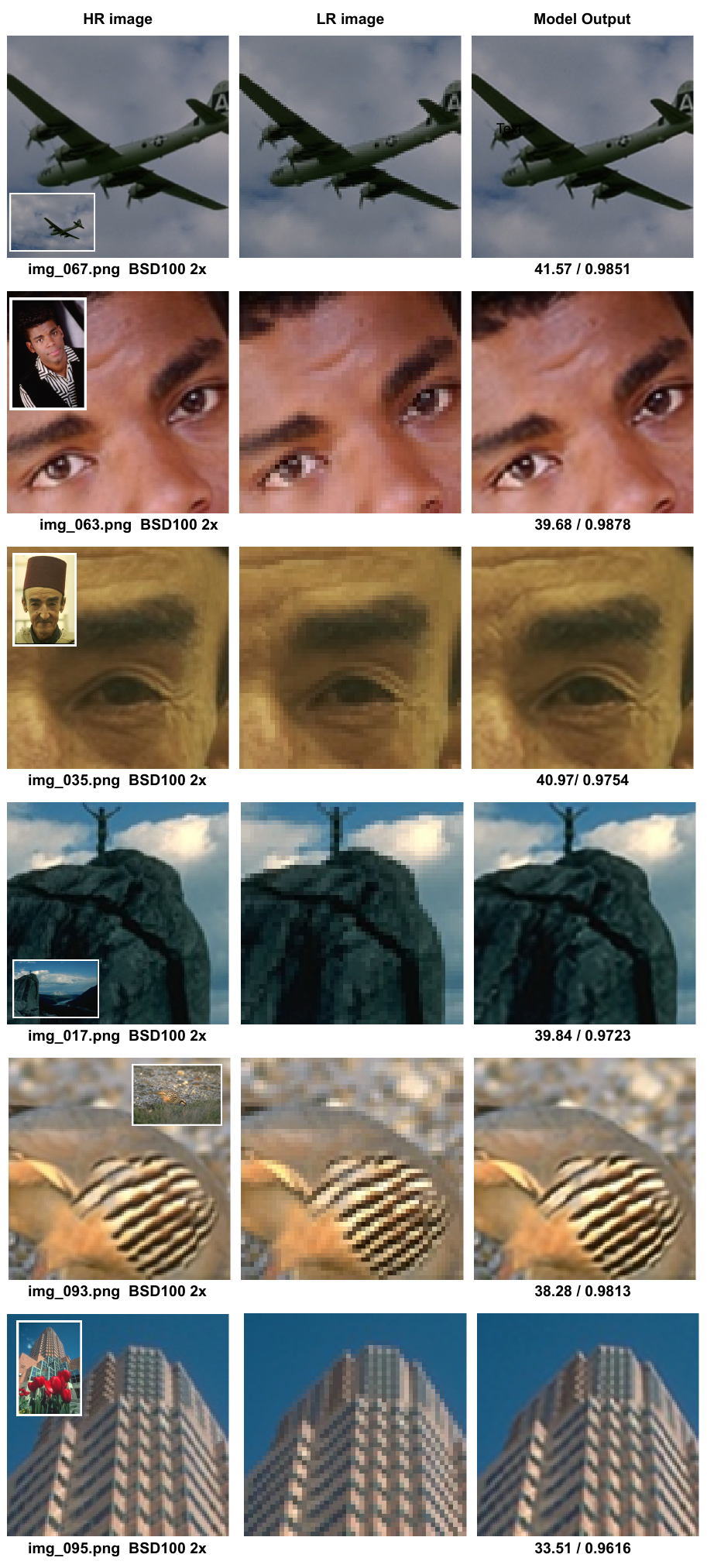}
\caption{Visual results of $2\times$ SR on BSD100 dataset. Each column from the left shows a patch from the HR image (shown as a small image near the corner), the same patch extracted from the LR image, and a patch taken from the model output respectively. The filename of the image is given below the HR image and the PSNR/SSIM of the model output is reported at below the model output. The values displayed are computed for the whole image and not just the patch.}
\label{fig:images}
\end{figure}

\section{Experiments and Results}

\subsection{Datasets}

We used DIV2K dataset~\cite{8014884} for training WaveMixSR. We did not employ any pre-training on larger datasets such as DF2K~\cite{8014884} or ImageNet~\cite{5206848} to compare the performance in training data-constrained settings. The performance of WaveMixSR was tested on four benchmark datasets -- BSD100~\cite{937655}, Urban100~\cite{7299156}, Set5~\cite{BMVC.26.135}, and Set14~\cite{10.1007/978-3-642-27413-8_47}, .

\begin{table}[]
\centering
\begin{tabular}{@{}lrr@{}}
\toprule
Model & \#Params. & \#Mutli-Adds. \\ \midrule
SwinIR~\cite{liang2021swinir} & 11.8 M & 49.6 G \\
HAT~\cite{chen2023activating} & 20.8 M & 103.7 G \\
\textbf{WaveMixSR} & \textbf{1.7 M} & \textbf{25.8 G} \\ \bottomrule
\end{tabular}
\vspace{2mm}
\caption{Model complexity comparison of WaveMixSR with other state-of-the-art methods such as SwinIR~\cite{liang2021swinir} and HAT~\cite{chen2023activating} on $4\times$ SR of $64\times64$ input patch.}
\label{tab:param}
\end{table}

\begin{table*}[t]
\centering
\begin{tabular}{@{}clllrrrr@{}}
\toprule
\multicolumn{1}{c}{\multirow{2}{*}{Scale}} & \multicolumn{1}{c}{\multirow{2}{*}{\begin{tabular}[c]{@{}c@{}}Training dataset\end{tabular}}} & \multicolumn{2}{c}{Set5} & \multicolumn{2}{c}{Set14} & \multicolumn{2}{c}{Urban100} \\ \cmidrule(l){3-8} 
\multicolumn{1}{c}{} & \multicolumn{1}{c}{} & \multicolumn{1}{c}{PSNR} & \multicolumn{1}{c}{SSIM} & \multicolumn{1}{c}{PSNR} & \multicolumn{1}{c}{SSIM} & \multicolumn{1}{c}{PSNR} & \multicolumn{1}{c}{SSIM} \\ \midrule
$\times2$ & DIV2K & 35.80 & 0.9543 & 31.27 & 0.9044 & 29.14 & 0.9078 \\
$\times3$ & DIV2K & 31.92 & 0.9145 & 28.77 & 0.8413 & 25.82 & 0.8193 \\
$\times4$ & DIV2K & 29.37 & 0.8627 & 26.25 & 0.7511 & 23.57 & 0.7300 \\ \bottomrule
\end{tabular}
\caption{Quantitative results of WaveMixSR on other benchmark SR datasets}
\label{tab:results}
\end{table*}

\subsection{Implementation Details}
All experiments were done with a single 80 GB Nvidia A100 GPU. We used AdamW optimizer ($\alpha = 0.001, \beta_{1} = 0.9, \beta_{2}=0.999, \epsilon = 10^{-8}$) with a weight decay of 0.01 during initial epochs and then used SGD with a learning rate of $0.001$ and momentum $= 0.9$ during the final 50 epochs~\cite{keskar2017improving, https://doi.org/10.48550/arxiv.2201.10271}. A dropout of 0.3 is used in our experiments. A batch size of 1 was used when the full-resolution images were passed to the model and a batch size of 432 was used when images were passed as $64\times64$ resolution patches.

The low-resolution images were generated from the HR images by using bicubic down-sampling in Pytorch. We used the full-resolution HR image as the target and generated the input LR image using down-sampling for each of the SR tasks. No data augmentations were used while training the WaveMixSR models. Huber loss was used to optimize the parameters. We used automatic mixed precision in PyTorch during training. For the quantitative results, PSNR and SSIM (calculated on the Y channel) are reported. 

The embedding dimension of 144 was used in WaveMix blocks. 4 WaveMix blocks were connected in series in the Y channel path. The convolutions layers before and after the WaveMix blocks which were used to vary channel dimensions employed $3\times3$ kernels with stride and padding set to 1 to maintain the feature resolution.

\subsection{Results}

Table~\ref{tab:bsd100} shows the quantitative performance comparison of WaveMixSR with other models such as EDSR~\cite{lim2017enhanced}, RCAN, SAN, IGNN, HAN, NLSN, RCAN-it, EDT, SwinIR~\cite{liang2021swinir}, SwinFIR and HAT on BSD100 dataset. 
We can see that WaveMixSR significantly outperforms all the other models. On BSD100 for $2\times$ SR, the performance is 0.34 dB greater than the prior state-of-the-art HAT-L. It also outperforms HAT-L by 0.0256 for $2\times$ SR and 0.0054 for $4\times$ SR on BSD100 in terms of the SSIM metric.

Furthermore, WaveMixSR achieves these performance gains without needing ImageNet pre-training like HAT-L and SwinFIR or training on a much larger DF2K dataset. BSD100 is a classical dataset composed of a large variety of images ranging from natural images to object-specific such as plants, people, and scenery. Unlike other datasets like Set5 and Set14 which has a low number of images (5 and 14 respectively), BSD100 has 100 images which give a more accurate metric of the model capability on common images. Urban100 has images of only urban settings with buildings, which mostly have a periodic lattice structure with windows and straight lines. BSD100 images have more variation in shapes since it contains a larger number of images of objects, scenery, animals, and people. All the images in BSD100 dataset are of $480\times320$ resolution. The performance of WaveMixSR on BSD100 
demonstrates our model's ability to generate natural HR images. 

The computational and parametric complexity of our method was compared with the other state-of-the-art transformer-based methods such as SwinIR and HAT as shown in Table~\ref{tab:param}. The number of Multiply-Add operations is counted at the input size of $64\times64$. We see that WaveMixSR is highly parameter-efficient compared to other models, due to the presence of parameter-free token-mixing using 2D-DWT. The self-attention, channel attention, and cross-attention in transformer-based methods introduce $20-10\times$ more parameters compared to WaveMixSR. WaveMixSR also saves on computing as it uses half of the multi-add operations compared to SwinIR and one-fourth that of HAT.

We provide the visual results of WaveMixSR in Figure~\ref{fig:images}. In ``img\_067", we can see that WaveMixSR was able to regenerate the star emblem on the side of the airplane and sharpen the image. On ``img\_063" and ``img\_035" we can see the ability of our model to regenerate facial features from LR images. Both eyes and eyebrows were reconstructed without blurring in ``img\_063". The wrinkles and texture of the face and  around the eyelids which were completely absent in the LR image were reconstructed successfully by our model. The blurry individual in ``img\_017" was fully reconstructed and sharpened along with the cliff rocks. WaveMixSR is also able to reconstruct stripes on the bird in ``img\_093" and recover lattice content such as windows on the sky-scrapper in ``img\_095".

We expect further improvement in the performance of our model when trained using the larger DF2K datasets and even with ImageNet pre-training as used in the previous state-of-the-art models. Use of adversarial training~\cite{ledig2017photorealistic} can also be tested for further improvements.

\subsection{Ablation Studies}

We observed that using bilinear interpolation and bicubic interpolation for upsampling the images in both the Y channel and CbCr channel path in WaveMixSR was giving similar SR performance. 

We experimented with various loss functions and their linear combinations to identify the most suitable loss function for SR. As shown in Table~\ref{tab:loss}, we found that Huber loss, which is the same as the L2 loss when the loss value is high and changes to L1 loss when the loss value becomes low, is most effective in optimizing  the model parameters.

We also experimented with a single path model where all three channels (YCbCr) are sent to the WaveMix blocks without having a separate path for CbCr channels. The SR results showed that sending CbCr channels does not increase performance. Experiments where the image was processed by WaveMix blocks in RGB space instead of YCbCr space also lowered the PSNR values by $\sim 3$dB. 

We varied the number of layers, embedding dimension of WaveMix blocks, levels of wavelet decomposition and dropout. Increasing the layers showed an increase in reconstruction quality initially but performance stagnated after four layers. The number of channels in WaveMix blocks (embedding dimension) too followed a similar pattern in performance after 144. We found that we do not need to use more than one level of 2D DWT for image super-resolution task as the receptive field expansion with just one level of Haar wavelet is enough for low-level vision tasks. Adding more levels did not improve the performance of the model. We also experimented with different values of dropout used inside the WaveMix blocks and found that a dropout of 0.3 gave an optimum performance.

\begin{table}[]
\centering
\begin{tabular}{@{}lll@{}}
\toprule
Loss Functions & PSNR & SSIM\\ \midrule
L1 Loss & 33.18 & 0.9197\\
L2 Loss & 32.24 & 0.9132\\
SSIM Loss & 32.11 & 0.9115\\
Charbonier Loss & 33.29 & 0.9204 \\
Huber Loss & \textbf{33.66} & \textbf{0.9256}\\ \bottomrule
\end{tabular}
\vspace{2mm}
\caption{Experiments on Div2k dataset with different loss functions shows that minimizing Huber Loss provides the best quantitative results}
\label{tab:loss}
\end{table}

\section{Conclusion}

In this paper, we propose WaveMixSR, a wavelet-based token-mixing model for image super-resolution. Our model uses 2D-DWT as a token-mixing operation for high-resolution image reconstruction. Experiments show that WaveMixSR performs well with natural images and achieves state-of-the-art results in the BSD100 dataset. It consumes only a fraction of the computational resources and is highly parameter efficient compared to other state-of-the-art models. Our model was able to outperform other models without resorting to pre-training on large datasets or adversarial training\cite{ledig2017photorealistic}. Our work shows that token-mixers can be employed in low-level vision tasks such as image reconstruction and can outperform transformer-based models using less resources.

\bibliographystyle{unsrt}  
\bibliography{references}

\end{document}